\documentclass[runningheads]{llncs}
\usepackage{comment}
\usepackage{graphicx}
\usepackage{multicol}
\usepackage{multirow}
\usepackage{booktabs}
\usepackage[colorlinks,
            linkcolor=blue,
            anchorcolor=black,
            citecolor=blue
            ]{hyperref}
\usepackage{tabularray}
\usepackage[table]{xcolor}
\usepackage{ulem}
\usepackage{bbding}
\usepackage{bm}
\usepackage{amsfonts}
\usepackage[marginal]{footmisc}

\begin{document}
\title{Integrating Human Parsing and Pose Network for Human Action Recognition}

\def\CICAISubNumber{172}  % Insert your submission number here
\begin{comment}
\titlerunning{CICAI2023 submission ID \CICAISubNumber} 
\authorrunning{CICAI2023 submission ID \CICAISubNumber} 
\author{Anonymous CICAI submission}
\institute{Paper ID \CICAISubNumber}
\end{comment}
%******************

% \begin{comment}
\titlerunning{IPP-Net}
\author{Runwei Ding\inst{1} \and
Yuhang Wen\inst{2} \and
Jinfu Liu\inst{2} \and
Nan Dai\inst{3} \and
Fanyang Meng\inst{4} \and
Mengyuan Liu\inst{1(}\Envelope\inst{)}
}
\authorrunning{R. Ding \& Y. Wen et al.}
\institute{Shenzhen Graduate School, Peking University, Shenzhen, China \and
Sun Yat-sen University, Shenzhen, China \and
Changchun University of Science and Technology, Changchun, China \and
Peng Cheng Laboratory, Shenzhen, China\\
\email{nkliuyifang@gmail.com} 
}
% \end{comment}
%******************
\maketitle              % typeset the header of the contribution
\footnote{R. Ding and Y. Wen—These authors contributed equally to this work.\\
This work was supported by the Basic and Applied Basic Research Foundation of Guangdong (No. 2020A1515110370) and the National Natural Science Foundation of China (No. 62203476).}
\vspace{-2em}
\begin{abstract}
Human skeletons and RGB sequences are both widely-adopted input modalities for human action recognition. However, skeletons lack appearance features and color data suffer large amount of irrelevant depiction. To address this, we introduce human parsing feature map as a novel modality, since it can selectively retain spatiotemporal features of the body parts, while filtering out noises regarding outfits, backgrounds, etc. We propose an Integrating Human Parsing and Pose Network (IPP-Net) for action recognition, which is the first to leverage both skeletons and human parsing feature maps in dual-branch approach. The human pose branch feeds compact skeletal representations of different modalities in graph convolutional network to model pose features. In human parsing branch, multi-frame body-part parsing features are extracted with human detector and parser, which is later learnt using a convolutional backbone. A late ensemble of two branches is adopted to get final predictions, considering both robust keypoints and rich semantic body-part features. Extensive experiments on NTU RGB+D and NTU RGB+D 120 benchmarks consistently verify the effectiveness of the proposed IPP-Net, which outperforms the existing action recognition methods. Our code is publicly available at \url{https://github.com/liujf69/IPP-Net-Parsing}.

\end{abstract}
\section{Introduction}
Human action recognition is an important task in the field of computer vision, which also has great research value and broad application prospects in human-robot interaction \cite{wen2023interactive} and visual media \cite{8247231,zheng2017image}. Most action recognition methods take human skeletons sequences \cite{zhang2022zoom,liu2022generalized,tu2022joint,liu2023temporal,liu2023novel} or color images \cite{tu2019action} as the input modality for the following reasons. Intuitively a human skeleton can be viewed as a natural topological graph, which can well represent body movements and is highly robust to environmental changes, thereby adopted in many studies using graph convolutional neural networks (GCNs)~\cite{tu2022joint,gsgcn2022,shiftgcn2020,dynamicgcn2020,CTR-GCN2021,Chi_2022_CVPR}. On the other hand, color images (RGB data) have rich appearance information, therefore some prior studies in action recognition employ Convolutional Neural Networks (CNNs)~\cite{vpn2020,tu2019action} to extract spatiotemporal features from RGB video frames. In recent years, methods \cite{vpn2020} of integrating skeletal and RGB data have emerged to make effective use of multimodal features for better action recognition.

\begin{figure}[t]
\includegraphics[width=\textwidth]{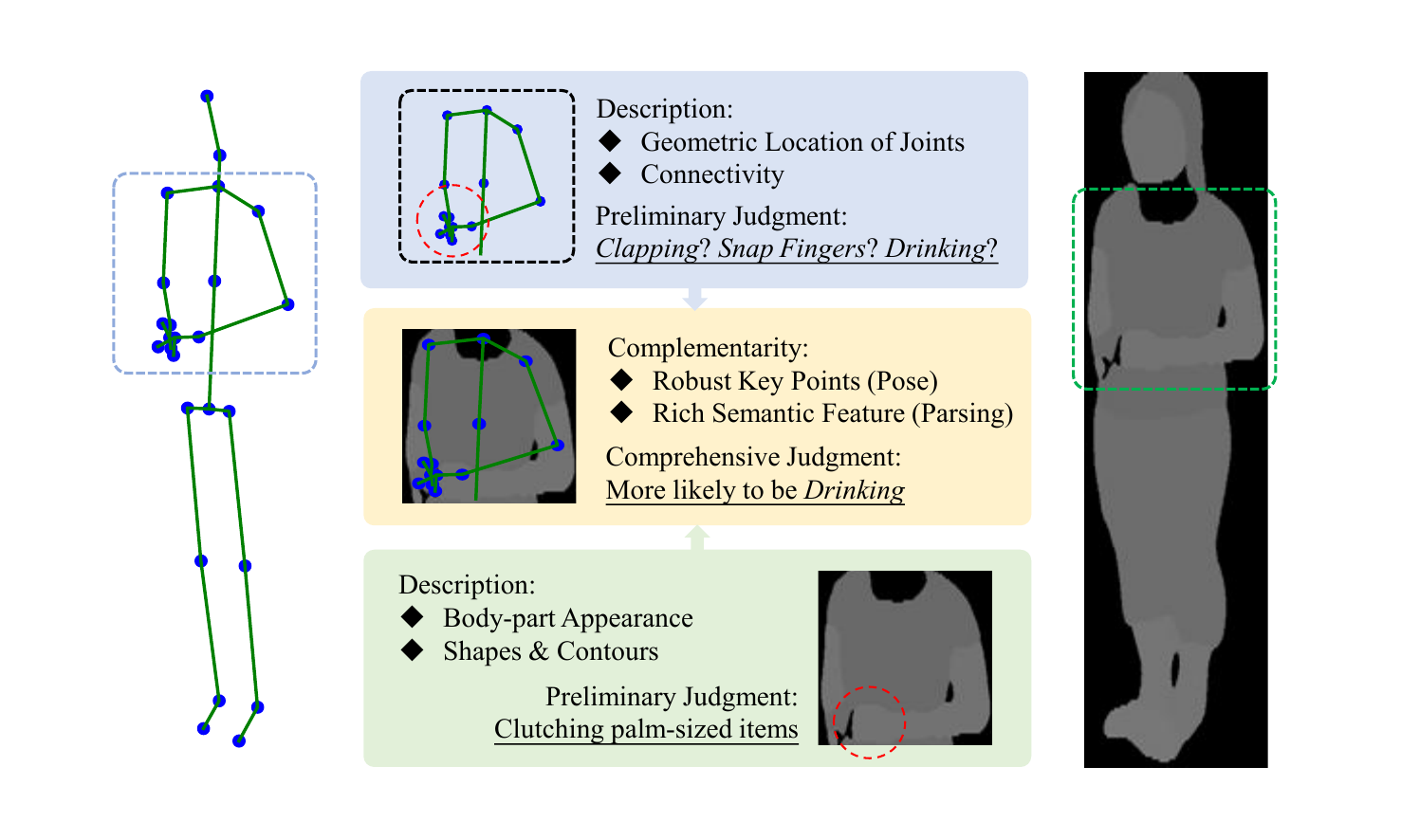}
\vspace{-2em}
\caption{An example of the complementarity between the human pose modality and the human parsing modality. A more comprehensive prediction can be made by integrating robust key points from poses and rich semantic features from parsing feature maps.} \label{teaser}
\end{figure}

However, both skeletal and RGB data have their respective limitations in representing human actions. Skeletons lack the ability to depict the appearance of human body parts, while RGB features are prone to being influenced by various sources of noise, including background interference, irrelevant elements, and changes in illumination. This naturally leads to the question: \uline{Can we explore a new modality that incorporates body-part appearance depiction while remaining noiseless and robust?}

Our affirmative response is inspired by the human parsing task, a vision task that holds significant importance in video surveillance and human behavior analysis. This task facilitates the recognition of distinct semantic parts, such as arms and legs, within the human body \cite{LIP2019}. By focusing on these semantic parts, human parsing explicitly and effectively eliminates action-irrelevant details, while retaining crucial extrinsic features of the human body. We believe this can effectively complement the skeletal data, as illustrated in Figure \ref{teaser}.

In this work, we advocate to integrate human parsing feature map as a novel modality into action recognition framework, and propose an Integrating Human Parsing and Pose Network (IPP-Net). Specifically, our IPP-Net consists of two trunk branches. In the first branch, referred to as the human pose branch, pose data is transformed into different skeleton representations and then fed into a graph convolutional neural network to obtain predictions. In the second branch, dubbed the human parsing branch, the human parsing features from multiple frames are sequentially combined to construct a feature map. This feature map is subsequently inputted into a convolutional neural network to derive recognition results. The results from both branches are integrated via a softmax layer to make final predictions. By leveraging the proposed IPP-Net, we effectively integrate pose data and human parsing feature maps to achieve better human action recognition.

The contributions of our work are summarized as follows:
\begin{enumerate}
    \item We advocate to leverage human parsing feature map as a new modality for human action recognition task, which is appearance-oriented depictive and also action-relevant.
    \item We propose a framework called Integrating Human Parsing and Pose Network (IPP-Net), which is the first to effectively integrates human parsing feature maps and pose data for robust human action recognition. Specifically, pose feature (representing body-part positions and connections) and human parsing feature (representing body-part contours and appearance) are learnt in two-stream approach and integrated via a late ensemble, to give comprehensive judgements about actions.
    \item Extensive experiments on benchmark NTU RGB+D and NTU RGB+D 120 datasets verify the effectiveness of our IPP-Net, which outperforms most existing action recognition methods.
\end{enumerate}

\section{Related Work}

\subsection{Human Action Recognition}
Prior approaches for human action recognition usually leverage skeleton data, which is a compact and sufficient representation for human actions. A significant body of work deals with more effective and efficient model architecture design for skeleton sequences \cite{dstanet2020,shiftgcn2020,dynamicgcn2020,CTR-GCN2021,Chi_2022_CVPR,gsgcn2022,psumnet2023,STSAnet2023,wen2023interactive}, with the aim of exploiting more informative joints. Beyond skeleton data, several of works use multi-modal features of human actions, including RGB sequences \cite{vpn2020} and text descriptions \cite{LST2022}, to achieve robust recognition results. VPN \cite{vpn2020} embeds 3D skeletons with their corresponding RGB videos, and feed them into an attention network to learn spatiotemporal relations. Empowered by Large Language Model (LLM), LST \cite{LST2022} conditions GCN training using the generated text descriptions of body-part motions. Compared to single modality, Multi-modal inputs provide unique information for each type of action, albeit in different forms, thereby enhancing human action understanding. However, it may contain irrelevant features (e.g. specific outfits, backgrounds). Our IPP-Net also leverages multi-modalities, and we argues to utilize pose data and feature map sequences of human parsing. Compared with the above approaches, human parsing filters out irrelevant information regarding illumination and backgrounds, while selectively retaining spatiotemporal features of all body parts.

\subsection{Human Parsing}
Human parsing involves the segmentation of a human image into fine-grained semantic parts, including the head, torso, arms, and legs. Several benchmarks have been proposed for human parsing task, providing large-scale annotations of body parts \cite{ATR,LIP2019}. A number of works concentrated on this problem and proposed novel models for better semantic parsing. The majority are based on ResNet architecture \cite{LIP2019,ruan2019devil,Zhao_2017_CVPR,li2020self}, while some are based on 
% HRNet architecture \cite{10.1007/978-3-030-58539-6_11} and 
Transformer architecture \cite{Chen_2023_CVPR}. Inspired by the human parsing task, we exploit the advantages of human parsing to get noiseless and concise representations, thus proposing a framework to ensemble parsing results and skeletons.

\begin{figure}[t]
\includegraphics[width=\textwidth]{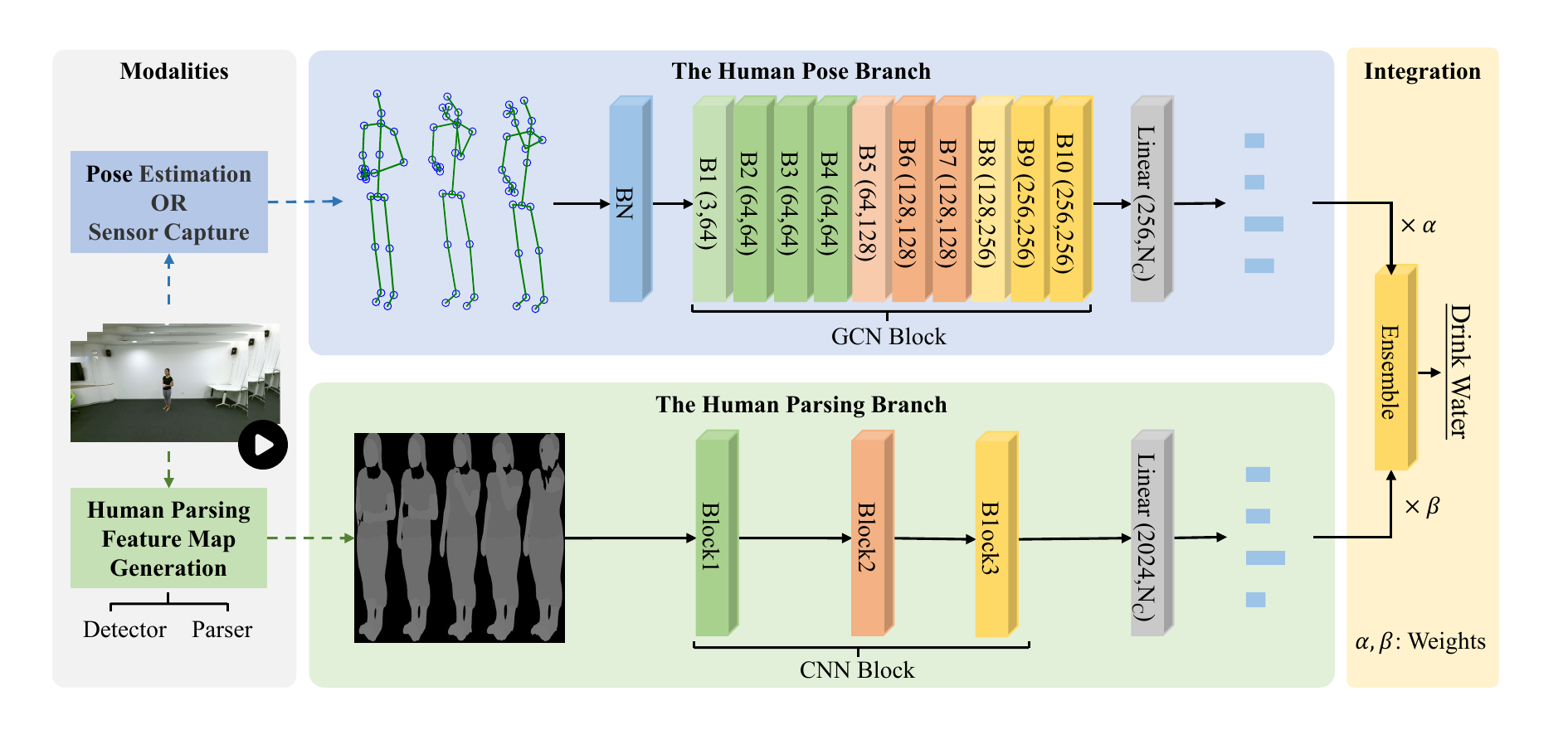}
\caption{Framework of our proposed Integrating Human Parsing and Pose Network.} \label{network}
\end{figure}

\section{IPP-Net}
As shown in Fig.~\ref{network}, in our proposed Integrating Human Parsing and Pose Network, we incorporate two primary branches, specifically the human pose branch and the human parsing branch. The human pose branch (Section \ref{section:Pose}) utilizes GCN to model the skeleton data, while the human parsing branch (Section \ref{section:Parsing}) employs CNN to extract deep features of the human parsing feature maps. Subsequently, the outcomes from these two branches are integrated via a late ensemble to get action predictions (Section \ref{section:Integration}). 

\subsection{Human Pose Learning}
\label{section:Pose}
\textbf{Skeleton Data.} Skeleton data is widely-used in action recognition frameworks. The pose branch of proposed IPP-Net leverages human skeleton data collected by sensors. Conceptually the skeleton sequence is a natural topological graph, in which joints are graph vertices, and bones are edges. The graph is denoted as $\textit{\textbf{G}} = \left\{\textit{\textbf{V}}, \textit{\textbf{E}} \right\}$, where $\textit{\textbf{V}} = \left\{\textit{\textbf{v}}_{1}, \textit{\textbf{v}}_{2}, \cdots, \textit{\textbf{v}}_{N} \right\}$ is a set of N joints and \textit{\textbf{E}} is a set of bones in the skeleton. For 3D skeleton data, the joint $\textit{\textbf{v}}_{i}$ is denoted as $\left\{\textit{\textbf{x}}_{i}, \textit{\textbf{y}}_{i}, \textit{\textbf{z}}_{i} \right\}$, where $\textit{\textbf{x}}_{i}$, $\textit{\textbf{y}}_{i}$ and $\textit{\textbf{z}}_{i}$ locate $\textit{\textbf{v}}_{i}$ in three-dimensional Euclidean space.

Skeleton data can be defined as four different modalities, namely \textit{joint} $(\textit{\textbf{J}})$, \textit{bone} $(\textit{\textbf{B}})$, \textit{joint motion} $(\textit{\textbf{JM}})$ and \textit{bone motion} $(\textit{\textbf{BM}})$. Given two joints data $\textit{\textbf{v}}_{i} = \left\{\textit{\textbf{x}}_{i}, \textit{\textbf{y}}_{i}, \textit{\textbf{z}}_{i} \right\}$ and $\textit{\textbf{v}}_{j} = \left\{\textit{\textbf{x}}_{j}, \textit{\textbf{y}}_{j}, \textit{\textbf{z}}_{j} \right\}$, a bone data of the skeleton is defined as a vector $\textit{\textbf{e}}_{\textit{\textbf{v}}_{i}, \textit{\textbf{v}}_{j}} = \left(\textit{\textbf{x}}_{i}-\textit{\textbf{x}}_{j}, \textit{\textbf{y}}_{i}-\textit{\textbf{y}}_{j}, \textit{\textbf{z}}_{i}-\textit{\textbf{z}}_{j} \right)$. Given two joints data $\textit{\textbf{v}}_{ti}$, $\textit{\textbf{v}}_{(t+1)i}$ from two consecutive frames, the data of joint motion is defined as $\textit{\textbf{m}}_{ti} = \textit{\textbf{v}}_{(t+1)i} - \textit{\textbf{v}}_{ti}$. Similarly, given two bones data $\textit{\textbf{e}}_{\textit{\textbf{v}}_{(t+1)i},\textit{\textbf{v}}_{(t+1)j}}$, $\textit{\textbf{e}}_{\textit{\textbf{v}}_{ti},\textit{\textbf{v}}_{tj}}$ from two consecutive frames, the data of bone motion is defined as $\textit{\textbf{m}}_{\textit{\textbf{v}}_{ti},\textit{\textbf{v}}_{tj}} = \textit{\textbf{e}}_{\textit{\textbf{v}}_{(t+1)i},\textit{\textbf{v}}_{(t+1)j}} - \textit{\textbf{e}}_{\textit{\textbf{v}}_{ti},\textit{\textbf{v}}_{tj}}$.

\textbf{Backbone.} GCN-based methods have dominated the skeleton-based action recognition task due to their unique advantages in modeling graph-structured data. Our IPP-Net also embraces GCN as backbone to learn pose features. A GCN is typically composed of graph convolutions and temporal convolutions. The normal graph convolution utilizes the weight \textit{\textbf{W}} for aggregate features of neighbor vertices through adjacency matrix $\textit{\textbf{a}}_{ij}$ to update vertice $\textit{\textbf{v}}_{i}$’s features $\textit{\textbf{f}}_{i}$, which is formulated as:
\begin{equation}
\textit{\textbf{f}}_{i} = \sum_{\textit{\textbf{v}}_{j} \in N(\textit{\textbf{v}}_{i})}\textit{\textbf{a}}_{ij}\textit{\textbf{x}}_{j}\textit{\textbf{W}},
\label{con:Equation1}
\end{equation}
where $\textit{\textbf{a}}_{ij}$ can be obtained in static (defined manually) or dynamic (initialized manually but learnable) ways. Our IPP-Net feeds the normalized four skeleton modalities defined above into the ten dynamic GCN blocks for feature extraction, followed by a linear layer to obtain the recognition result.

\subsection{Human Parsing Learning}
\label{section:Parsing}
\begin{figure}[t]
\includegraphics[width=\textwidth]{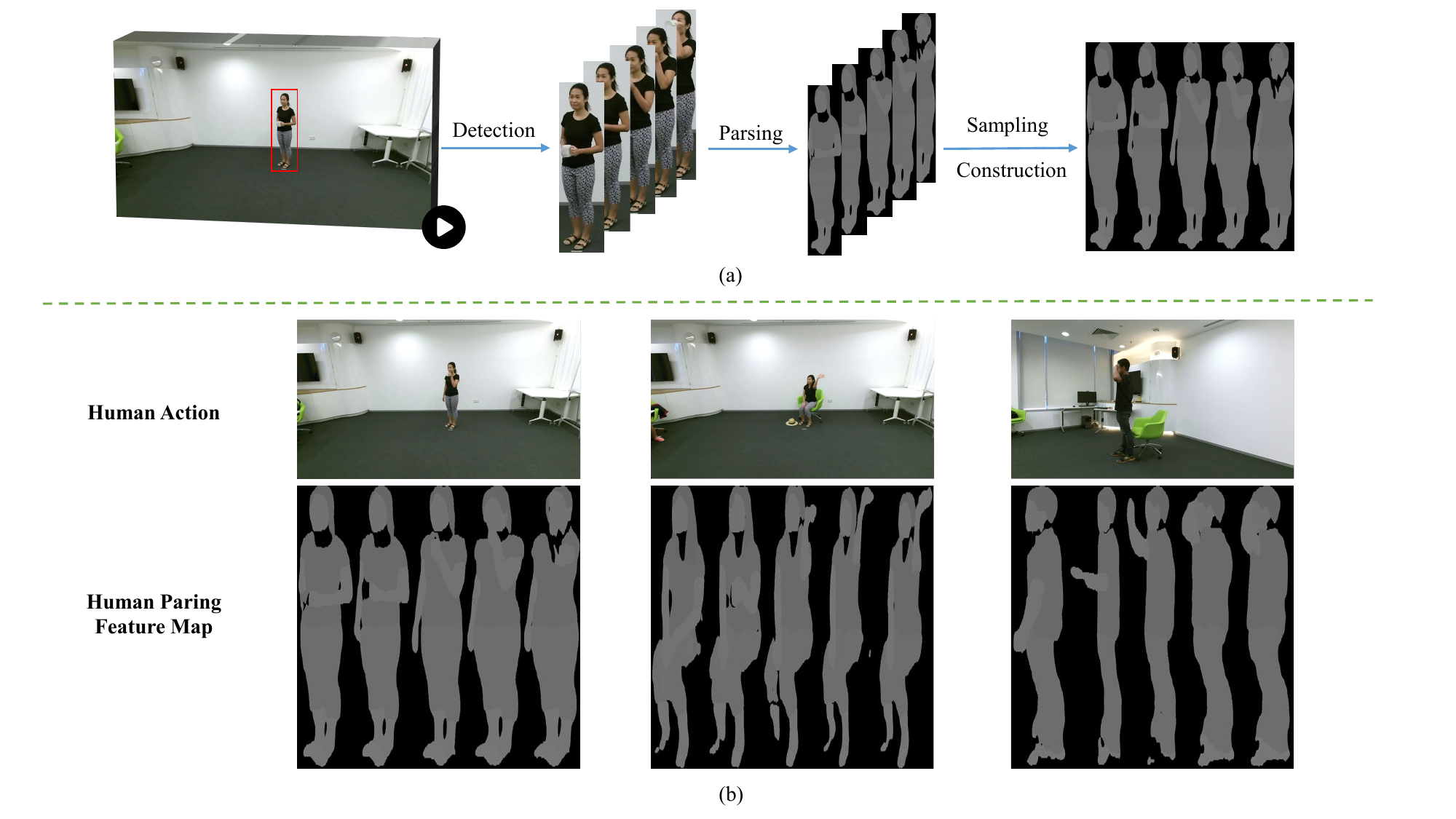}
\vspace{-1em}
\caption{(a) Illustration of human parsing feature map generation from RGB videos. (b) Visualization of human parsing feature maps for different action samples.}  \label{parsing}
\end{figure}

\textbf{Human Parsing Feature Map Generation.} Inspired by the human parsing task, we propose human parsing feature maps as one novel modality representing body-part movements with appearances. The following steps shows how we generate this new modality from raw RGB videos. Given an RGB video stream $\textit{\textbf{F}} = \left\{\textit{\textbf{f}}_{1}, \textit{\textbf{f}}_{2}, \cdots, \textit{\textbf{f}}_{N} \right\}$ with $N$ frames, frame-level feature is extracted by
\begin{equation}
\textit{$\textit{\textbf{I}}_{i}$} = E(D(\textit{\textbf{f}}_{i})),
\label{con:Equation2}
\end{equation}
where $1\leq i\leq N, i \in \mathbb{N}$, then $D$ and $E$ denote an object detector and a feature extractor respectively. In implementation, we use YoloV5~\cite{yolov5} as the target detector of our model, and feed the detected human maps into Resnet101~\cite{he2015deep} to extract the features $\textit{\textbf{I}}_{i}$ of each frame. 

The frame-level feature $\textit{\textbf{I}}_{i}$ gets downsampled and upsampled in PSPNet~\cite{Zhao_2017_CVPR} for human parsing, formulated as
\begin{equation}
\bm{\hat{P}}_{i} = \mathrm{argmax}_{class}(PSPNet(\textit{\textbf{I}}_{i})),
\label{con:Equation3}
\end{equation}
where $1\leq i\leq N, i \in \mathbb{N}$. $\bm{\hat{P}}_{i}$ retrieves the index of the most possible body part category for each pixel. 

Then the frame-level maps get resized to a standard shape [h, w]. We sample total $N_{sample}$ frames from $N$ frame-level maps with the random equidistant sampling strategy, denoted as 
\begin{equation}
    \bm{\tilde{P}}_{j} = \bm{\hat{P}}_{1+\delta\times (j-1)} 
\end{equation}
where $1\leq j\leq N_{sample}, j \in \mathbb{N}$. Suppose $N_{sample} \leq N$, then $\delta$ is a random positive integer between 1 and $\lfloor N/N_{sample} \rfloor$. If $N_{sample} > N$, then we repeat the $N$ maps until $N_{sample}$ frames are sampled. The chosen $N_{sample}$ feature maps are chronologically arranged to construct the final feature map $\bm{P}$. Fig.~\ref{parsing} (b) visualizes the feature maps of three action samples, namely \textit{drink water}, \textit{hand waving} and \textit{salute}.

\textbf{Backbone.} Intuitively the human parsing feature maps repeated in channel dimension can be viewed as 3-channel grayscale images. Therefore, a convolutional backbone is adopted to extract deep parsing features for its strong perception for locality, which is essential to percept graphic structures, such as edges and shapes in parsing feature maps $\bm{P}$.

\subsection{Integration}
\label{section:Integration}
The outcomes from the human pose branch and the human parsing branch are fused via an ensemble layer, which is formulated as:
\begin{equation}
\textit{$\textit{\textbf{S}}$} = \mathrm{softmax}(\alpha \cdot C(\textit{\textbf{V}}) + \beta \cdot M(\textit{\textbf{P}})),
\label{con:Equation4}
\end{equation}
where \textit{\textbf{V}} and \textit{\textbf{P}} denote the skeleton data and human parsing feature maps respectively. The skeleton data \textit{\textbf{V}} and feature maps \textit{\textbf{P}} are respectively fed into a graph convolutional neural network $C$ and a convolutional neural network $M$. The parameters $\alpha$ and $\beta$ represent the ensemble weights for the linear combination. After a softmax layer, we could obtain the final prediction \textit{\textbf{S}}.

\section{Experiments}
\subsection{Datasets}
Experiments are conducted on two widely-used large-scale human action recognition datasets, illustrated as follows:

\textbf{NTU-RGB+D} \cite{7780484}, also referred as \textbf{NTU 60}, is a widely used 3D action recognition dataset containing 56,880 video samples. The action samples are performed by 40 distinct subjects and categorized into 60 classes. The original paper \cite{7780484} suggests two benchmark scenarios for evaluation: (1) Cross-View (X-View), where the training data originates from cameras at $0^\circ$ (view 2) and $45^\circ$ (view 3), and the testing data is sourced from the camera at $45^\circ$ (view 1). (2) Cross-Subject (X-Sub), where the training data comprises samples from 20 subjects, while the remaining 20 subjects are reserved for testing.

\textbf{NTU-RGB+D 120} \cite{8713892}, also referred as \textbf{NTU 120}, is derived from the NTU-RGB+D dataset. A total of 114,480 video samples across 120 classes are performed by 106 volunteers and captured using three Kinect V2 cameras. The original work \cite{8713892} also suggests two criteria: (1) Cross-subject (X-Sub), where the training data is sourced from 53 subjects, while the testing data originates from the other 53 subjects. (2) Cross-setup (X-Set), where the training data is composed of samples with even setup IDs, and the testing data comprises samples with odd ones.

\subsection{Implementation Details}
All experiments are conducted on four Tesla V100-PCIE-32GB GPUs and two NVIDIA GeForce RTX 3070 GPUs. On the pose branch, we adopt CTR-GCN \cite{CTR-GCN2021} as the backbone. We use SGD for model training, conducting 65 epochs with a batch size of 64. The initial learning rate is set to 0.1 and decayed by a factor of 0.1 at epochs 35 and 55. On the parsing branch, we adopt InceptionV3 \cite{szegedy2016rethinking} as the backbone. We also use SGD to train the model for 30 epochs with a batch size of 64. The initial learning rate is set to 0.1 and decayed by a factor of 0.0001 at epochs 10 and 25. The cross-entropy loss is employed as the training loss. During training we select 5 frames randomly to construct the feature maps of human parsing, while 5 frames at equal intervals in testing. When generating the human parsing feature map, we resize the human parsing map of each frame to the specified size [480, 96] and five frames of which will be arranged in chronological order to form the final feature map with a size of [480, 480].

% Sota Table
\begin{table}[t]
    \caption{Accuracy comparison with state-of-the-art methods on NTU-RGB+D and NTU-RGB+D 120 dataset.}
    \centering
    \label{tab:Example}
    % \resizebox{\textwidth}{!}{
    \begin{tblr}
        {
            colspec=clccccc,
            vline{2,3,4,5,6,7}, hline{1,2,3,13,17},
            cell{1}{1}={r=2}{c}, cell{1}{2}={r=2}{c},
            cell{1}{3}={r=2}{c}, cell{1}{4}={c=2}{c},
            cell{1}{6}={c=2}{c}, cell{3}{1}={r=10}{c},
            cell{13}{1}={r=4}{c}, row{12} = {valign=m},
            cell{3-Z}{1} = {cmd=\rotatebox{90}},
            cell{15}{3}={r=2}{c},
            row{3-12} = {yellow!10},
            row{13-Z} = {cyan!5},
        }
    \textbf{Type} & \textbf{Method} & \textbf{Source} & \textbf{NTU 60} ($\%$) & & \textbf{NTU 120} ($\%$) & \\
    & & & \textbf{X-Sub} & \textbf{X-View} & \textbf{X-Sub} & \textbf{X-Set}\\
    Pose & Shift-GCN \cite{shiftgcn2020} & CVPR'20 & 90.7 & 96.5 & 85.9 & 87.6 \\
    & DynamicGCN \cite{dynamicgcn2020} & MM'20 & 91.5 & 96.0 & 87.3 & 88.6 \\
    & DSTA-Net \cite{dstanet2020} & ACCV'20 & 91.5 & 96.4 & 86.6 & 89.0 \\
    & MS-G3D \cite{MS-G3D2020} & CVPR'20 & 91.5 & 96.2 & 86.9 & 88.4 \\
    & MST-GCN \cite{chen2021multi} & AAAI'21 & 91.5 & 96.6 & 87.5 & 88.8 \\
    & CTR-GCN \cite{CTR-GCN2021} & ICCV'21 & 92.4 & 96.8 & 88.9 & 90.6 \\
    & GS-GCN \cite{gsgcn2022} & CICAI'22 & 90.2 & 95.2 & 84.9 & 87.1 \\
    & PSUMNet \cite{psumnet2023} & ECCV'22 & 92.9 & 96.7 & 89.4 & 90.6 \\
    & InfoGCN \cite{Chi_2022_CVPR} & CVPR'22 & 93.0 & 97.1 & 89.8 & 91.2 \\
    & STSA-Net \cite{STSAnet2023} & {Neurocom\\puting'23} & 92.7& 96.7 & 88.5 & 90.7 \\
    Multi-Modality& VPN \cite{vpn2020} & ECCV'20 & 93.5 & 96.2 & 86.3 & 87.8 \\
    & LST \cite{LST2022} & arXiv'22 & 92.9 & 97.0 & 89.9 & 91.1 \\
    % & Ours (J Only) &  & 90.2 & 95.0 & 85.0 & 86.7 \\
    & Ours (J+B+P) &  & 93.4 & 96.8 & 89.4 & 91.2 \\
    & \textbf{Ours} &  & \textbf{93.8} & \textbf{97.1} & \textbf{90.0} & \textbf{91.7} \\
    \end{tblr}
    % }
    \label{tab:sota}
\end{table}

\subsection{Comparison with Related Methods}
In Table \ref{tab:sota}, we compare our IPP-Net with the existing human action recognition methods on the NTU-RGB+D and NTU-RGB+D 120 datasets. In these two large-scale action recognition datasets, our model outperforms all existing methods under nearly all evaluation benchmarks. Notably, our IPP-Net is the first one that combines human parsing and pose data for action recognition. 

\begin{figure}[t]
    \includegraphics[width=\textwidth]{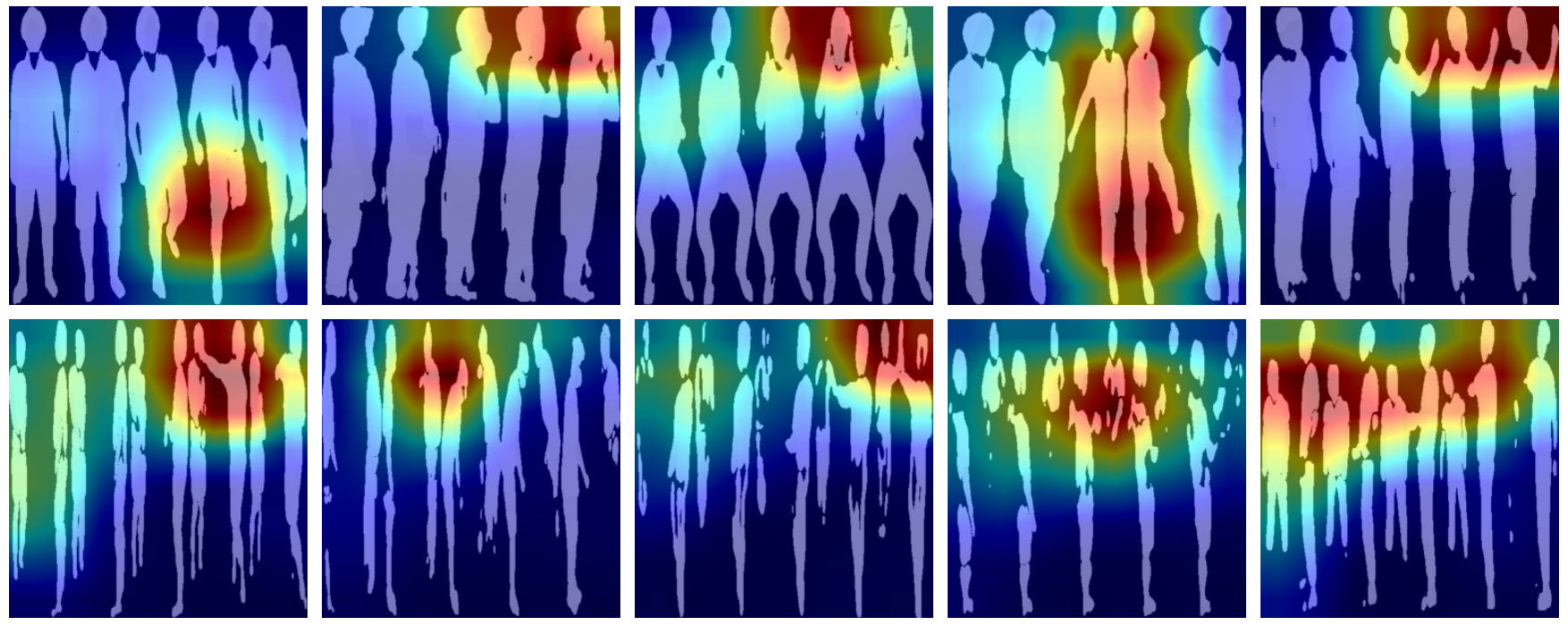}
    \caption{Visual explanation with class activation maps on how the convolutional backbone judges the human action based on the given parsing feature maps.} 
    \label{viz}
\end{figure}

In the NTU-RGB+D dataset, the top-1 accuracy is 93.8$\%$ and 97.1$\%$ on the benchmark of X-Sub and X-View respectively, which outperforms CTR-GCN \cite{CTR-GCN2021} by 1.4$\%$ and 0.3$\%$. On the tougher benchmark namely X-Sub, our IPP-Net outperforms LST \cite{LST2022} by 0.9$\%$, even though the LST model additionally introduces texts as a new modality. In the NTU-RGB+D 120 dataset, the top-1 accuracy is 90.0$\%$ and 91.7$\%$ on the benchmark of X-Sub and X-View respectively, which outperforms CTR-GCN by 1.1$\%$ and 1.1$\%$. The most related method to our work is VPN \cite{vpn2020}, which introduces RGB features besides skeletons for human action recognition. On two benchmarks of NTU-RGB+D 120 dataset, our IPP-Net outperforms VPN by 3.7$\%$ and 3.9$\%$ respectively. Fig. \ref{viz} visually illustrates how human parsing feature maps in our IPP-Net help recognize actions by providing semantic information about body parts. The class activation maps indicates that the human parsing branch can focus on the most informative body parts.

% Modality Table
\begin{table}[t]
\caption{Accuracy of different modalities on NTU-RGB+D and NTU-RGB+D 120.}
\centering
\begin{tabular}{ccccc|c|c|c|c}
\hline
\multicolumn{5}{c|}{\textbf{Modality}} & \multicolumn{2}{c|}{\textbf{NTU 60} ($\%$)} & \multicolumn{2}{c}{\textbf{NTU 120} ($\%$)}\\\hline
J & B & JM & BM & P & \textbf{X-Sub} & \textbf{X-View} & \textbf{X-Sub} & \textbf{X-Set} \\\hline
\Checkmark & & & & & 90.2 & 95.0 & 85.0 & 86.7 \\
& \Checkmark & & & & 90.5 & 94.7 & 86.2 & 87.5 \\
& & \Checkmark & & & 88.1 & 93.2 & 81.2 & 83.0 \\
& & & \Checkmark & & 87.3 & 92.0 & 81.7 & 82.9 \\
& & & & \Checkmark & 73.5 & 74.2 & 53.6 & 65.8 \\\hline
\Checkmark & & & & \Checkmark & 91.7 & 95.9 & 86.1 & 88.8 \\
& \Checkmark & & & \Checkmark & 91.8 & 95.9 & 87.5 & 89.8 \\
\Checkmark & \Checkmark & & & & 92.2 & 96.2 & 88.7 & 90.2 \\
\Checkmark & \Checkmark & \Checkmark & \Checkmark & & 92.4 & 96.5 & 89.0 & 90.5 \\
\Checkmark & \Checkmark & & & \Checkmark & 93.4 & 96.8 & 89.4 & 91.2 \\\hline
\Checkmark & \Checkmark & \Checkmark & \Checkmark & \Checkmark & \textbf{93.8} & \textbf{97.1} & \textbf{90.0} & \textbf{91.7} \\\hline
\end{tabular}
\label{tab:Modality}
\end{table}

\subsection{Ablation Study}
\textbf{Modality.} Table \ref{tab:Modality} reports the recognition accuracy of four skeleton modalities and human parsing feature map in NTU60 and NTU120 datasets respectively. When the joint, bone and human parsing feature maps are integrated together (J+B+P), the ensemble recognition accuracy of two benchmarks in the NTU60 dataset is 93.4$\%$ and 96.8$\%$, which outperforms the setting that combining joint and bone (J+B) by 1.2$\%$ and 0.6$\%$ respectively. Similarly, the setting (J+B+P) outperforms the setting (J+B) by 0.7$\%$ and 1.0$\%$ on the two benchmarks of the NTU120 dataset respectively. It can conclude from the experimental results that integrating human parsing as a modality into the framework can improve action recognition performance, which is attributed to the action-relevant body-part appearance provided by human parsing feature maps.

% frame Table
\begin{table}[t]
\caption{Comparison of different numbers of frames in feature map construction.}
\centering
\begin{tabular}{c|c|c}
\hline
\textbf{$\#$Frame} & \textbf{Parsing} ($\%$) & \textbf{Ensemble} ($\%$) \\\hline
3 & 46.7 & 89.8 \\\hline
4 & 50.4 & 89.9 \\\hline
5 & 53.6 & \textbf{90.0} \\\hline
6 & 55.6 & 89.9 \\\hline
\end{tabular}
\label{tab:frame}
\end{table}

% CNN Table
\begin{table}[t]
\caption{Accuracy of different CNN backbones in human parsing branch.}
\centering
\begin{tabular}{c|c|c}
\hline
\textbf{Backbone} & \textbf{Parsing} ($\%$) & \textbf{Ensemble} ($\%$) \\\hline
VGG11 \cite{simonyan2015deep} & 49.0 & 89.8 \\\hline
VGG13 \cite{simonyan2015deep} & 48.7 & 89.8 \\\hline
ResNet18 \cite{he2015deep} & 50.5 & 90.0 \\\hline
InceptionV3 \cite{szegedy2016rethinking} & 53.6 & \textbf{90.0} \\\hline
\end{tabular}
\label{tab:CNN}
\end{table}

\textbf{Frames for human parsing feature maps.} We explore how numbers of frames for constructing human parsing feature maps affect the accuracy on NTU120 X-Sub benchmark. As depicted in Table \ref{tab:frame}, we use 3-frame, 4-frame, 5-frame and 6-frame settings to make up feature maps respectively. The shapes of the four feature maps are all generated as [480,480] and then resized to [299,299] to meet InceptionV3's demand. We observe that the ensemble accuracy is 90.0$\%$ when using 5-frame parsing maps, which is the highest among the four settings.

\textbf{Backbones to learn parsing features.} On NTU120 X-Sub benchmark, we implement four different CNN backbones to learn parsing features. Result in Table \ref{tab:CNN} shows that InceptionV3 gets the highest ensemble accuracy among all the CNN backbones.

\section{Conclusions}
This work proposes an Integrating Human Parsing and Pose Network (IPP-Net) for human action recognition, which introduces the human parsing feature maps as a new modality to represent human actions. Human parsing can selectively preserve spatiotemporal body-part features while filtering out action-irrelevant information in RGB data. As a multi-modal action recognition framework, our IPP-Net is the first to leverage both skeletons and human parsing feature maps, considering both robust keypoints and rich semantic body-part features. The effectiveness of IPP-Net is verified on the NTU-RGB+D and NTU-RGB+D 120 datasets, where our IPP-Net outperforms most existing methods.

\normalem
\bibliographystyle{splncs04}
\bibliography{reference}

\end{document}